# AI-based Prediction of Biochemical Recurrence from Biopsy and Prostatectomy Samples

## Author Information

Andrea Camilloni[1], Chiara Micoli[1], Nita Mulliqi[2], Erik Everett Palm[1], Thorgerdur Palsdottir[1], Kelvin Szolnoky[1], Xiaoyi Ji[1], Sol Erika Boman[1,3], Andrea Discacciati[1], Henrik Grönberg[1], Lars Egevad[4], Tobias Nordström[1,5], Kimmo Kartasalo[2], Martin Eklund[1]

## Affiliations

[1] Department of Medical Epidemiology and Biostatistics, Karolinska Institutet, Stockholm, Sweden

[2] Department of Medical Epidemiology and Biostatistics, SciLifeLab, Karolinska Institutet, Stockholm, Sweden

[3] Department of Molecular Medicine and Surgery, Karolinska Institutet, Stockholm, Sweden

[4] Department of Oncology and Pathology, Karolinska Institutet, Stockholm, Sweden

[5] Department of Clinical Sciences at Danderyd Hospital, Karolinska Institutet, Stockholm, Sweden

## Contributions

A.C. performed all analyses, developed the modelling framework, and drafted the manuscript.

M.E. conceived and supervised the study.

T.N. contributed to manuscript development and refinement.

C.M., T.P., and A.D. provided statistical expertise and supported data acquisition.

N.M. contributed to AI design and the high-performance computing framework.

N.M., K.S., X.J., E.E.P., S.E.B., K.K. contributed to software implementation and data curation.

K.K. contributed to AI design and the high-performance computing framework and provided computational resources.

L.E. contributed to pathological grading of the STHLM3 samples.

H.G. conceived and designed the STHLM3 study.

All authors reviewed the manuscript and approved the final version.

## Corresponding author

martin.eklund@ki.se
2

# Abstract

Biochemical recurrence (BCR) after radical prostatectomy (RP) is a surrogate marker for aggressive prostate cancer with adverse outcomes, yet current prognostic tools remain imprecise. We trained an AI-based model on diagnostic prostate biopsy slides from the STHLM3 cohort ($n_{patients}$ = 676) to predict patient-specific risk of BCR, using foundation models and attention-based multiple instance learning. Generalizability was assessed across three external RP cohorts: LEOPARD ($n_{patients}$ = 508), CHIMERA ($n_{patients}$ = 95), and TCGA-PRAD ($n_{patients}$ = 379). The image-based approach achieved 5-year time-dependent AUCs of 0.64, 0.70, and 0.70, respectively. Integrating clinical variables added complementary prognostic value and enabled statistically significant risk stratification. Compared with guideline-based CAPRA-S, AI incrementally improved postoperative prognostication. These findings suggest biopsy-trained histopathology AI can generalize across specimen types to support preoperative and postoperative decision making, but the added value of AI-based multimodal approaches over simpler predictive models should be critically scrutinized in further studies.

# Introduction

Biochemical recurrence (BCR) following radical prostatectomy (RP) occurs in up to 30% of patients with localized prostate cancer and is widely used as an early surrogate marker of aggressive disease biology, being associated with an increased risk of subsequent clinical progression and metastatic spread[1]. Identifying patients at high risk of recurrence early is therefore important for guiding postoperative management, including the potential use of adjuvant therapy or intensified follow-up strategies[1]. Further, if patients preoperatively can be identified as having a high risk of recurrence if undergoing surgery, that information can impact primary treatment plans.

Current pre-surgery risk stratification approaches rely on clinical and pathological features such as Gleason score, clinical stage, and prostate-specific antigen (PSA) levels, while postoperative stratification nomograms also include features like surgical margins[2]. While informative, these variables provide an incomplete representation of the underlying biological heterogeneity of prostate cancer. As a result, substantial variability in outcomes persists within clinically defined risk groups, highlighting the need for more precise prognostic tools capable of capturing latent markers of aggressive disease[3].

Recent advances in artificial intelligence (AI) have enabled the extraction of prognostic information directly from routine histopathology slides, offering a scalable and potentially more reproducible alternative to traditional prostate cancer risk classification systems[4–10]. While AI-based grading systems have demonstrated promising accuracy, they have up to this point focused primarily on reproducing existing diagnostic frameworks, such as Gleason grading[11–13]. More recent work has shown that AI models can leverage additional information embedded in whole-slide images (WSIs), including spatial and morphological features that go beyond Gleason patterns[4,14].

To date, only a limited number of studies have addressed the prediction of BCR risk directly from digitized histopathological samples[4,5,8]. Among these, only one study showed that prognostic information related to BCR can be extracted preoperatively from prostate biopsy samples[8].

Recently, the LEarning biOchemical Prostate cAncer Recurrence from histopathology sliDes (LEOPARD) Challenge 2024[15] and the Combining HIstology, Medical imaging (radiology) and molEcular data for medical pRognosis and diagnosis (CHIMERA) Challenge 2025[16] were organized to promote the development of postoperative AI models for BCR prediction from RP specimens. Here, we developed an AI model trained on preoperative diagnostic biopsy slides from the STHLM3[17] cohort and evaluated its generalizability across three external RP cohorts: LEOPARD, CHIMERA, and TCGA-PRAD. This multi-cohort evaluation assesses the potential of AI-based histopathology to provide early and robust prognostic insights in prostate cancer, with possible clinical applications at the time of diagnosis.

Our goal is to assess whether histopathology-based AI models can improve BCR risk stratification compared with standard clinical predictors, and to evaluate their potential clinical utility in both preoperative and postoperative settings.

## Results

We evaluated our AI model across four independent prostate cancer cohorts comprising both biopsy and prostatectomy specimens. The development cohort, STHLM3 ($n_{patients}$ = 676, $n_{BCR}$ = 183, $t_{median}$=9.0 years), included pre-treatment diagnostic biopsies from patients who subsequently underwent RP. For external validation, we used three publicly available prostatectomy cohorts: LEOPARD ($n_{patients}$ = 508, $n_{BCR}$ = 87, $t_{median}$=4.8 years), CHIMERA ($n_{patients}$ = 95, $n_{BCR}$ = 27, $t_{median}$=3.2 years), and TCGA-PRAD ($n_{patients}$ = 379, $n_{BCR}$ = 48, $t_{median}$=2.3 years). The 5-year BCR-free survival rates were 86% in both STHLM3 and LEOPARD, 70% in CHIMERA, and 81% in TCGA-PRAD, highlighting cohort-level differences in recurrence risk. A summary of cohort characteristics is provided in Table 1.

## Model development

We developed and evaluated three distinct prognostic modalities: (i) a clinical-only model, using preoperative clinical features, including age, pre-treatment PSA, and biopsy International Society of Urological Pathology (ISUP) grade group, included as a reference benchmark for AI development; (ii) an image-only AI model using tile-level embeddings extracted from histopathology WSIs; and (iii) a multimodal AI model combining image and clinical features.

For the image-only and the multimodal pipelines, WSIs were processed using three pretrained foundation models—UNI2[18], Virchow2[19], and CONCH[20]—to extract high-dimensional tile-level embeddings. Additionally, an ensemble representation was constructed by concatenating the embeddings from all three models. These tile-level features were aggregated into patient-level representations using an attention-based multiple instance learning (MIL) framework[21], which outputs a continuous risk score optimized with a Cox proportional hazards loss function.

Each modality—clinical-only, image-only, and multimodal—was trained separately on the STHLM3 cohort. The multimodal model integrated clinical variables and image-derived features prior to prediction, allowing the model to learn a unified risk score.

To assess prognostic performance, we evaluated all combinations of model modality (clinical-only, image-only, and multimodal), and for image-based and multimodal models, all foundation model encoders (UNI2, Virchow2, CONCH, and ensemble). Based on internal cross-validation results, we selected the best-performing model from each modality using 5-year AUC as the primary criterion. Specifically, we chose the baseline clinical-only model trained on age, pre-treatment PSA, and ISUP grade, the image-only model using CONCH embeddings, and the multimodal model combining clinical variables with UNI2-derived image features. These selected models achieved a 5-year AUC of $0.70 \pm 0.12$ (clinical-only), $0.70 \pm 0.07$ (CONCH - image-only), and $0.73 \pm 0.03$ (UNI2 - multimodal), respectively. Full internal validation results across all combinations are shown in Figure 1 and Supplementary Table S3.

These three model designs were subsequently used for external validation, risk stratification, and downstream analyses.

## Model validation

To assess generalizability, we retrained the best-performing models from each modality on the full STHLM3 cohort and evaluated them on three external RP cohorts. Overall, model performance varied across cohorts: the multimodal model showed the highest discrimination in CHIMERA, whereas the clinical-only model performed best in TCGA-PRAD, highlighting cohort-dependent differences in the relative contribution of image-derived and clinical features.

In the LEOPARD cohort, which included only image data, the image-only model achieved a 5-year AUC of 0.64 (95% CI: 0.55–0.72).

In the CHIMERA cohort, the clinical-only model achieved a 5-year AUC of 0.80 (95% CI: 0.64–0.92), while the image-only model showed lower discrimination (AUC 0.70, 95% CI: 0.54–0.84). The multimodal model demonstrated higher discrimination than both single-modality approaches, achieving a 5-year AUC of 0.82 (95% CI: 0.69–0.94).

In the TCGA-PRAD cohort, the clinical-only model yielded the highest discrimination (5-year AUC of 0.76, 95% CI: 0.66–0.85). The multimodal model achieved a 5-year AUC of 0.72 (95% CI: 0.61–0.82), while the image-only showed lower discrimination (AUC 0.70, 95% CI: 0.60–0.80). The multimodal model showed higher discrimination than the image-only approach while not exceeding the clinical model in this cohort.

Full results across all modalities and cohorts are shown in Figure 2, and comprehensive performance metrics across modalities and encoders are provided in Supplementary Table S4.

## Comparison with guideline-based risk stratification

To contextualize the prognostic performance of the multimodal AI-based model within current clinical practice, we compared it with the Cancer of the Prostate Risk Assessment post-Surgical (CAPRA-S) score[22], a guideline-supported tool for postoperative risk stratification following RP. CAPRA-S scores were calculated for all external RP cohorts with available clinicopathological variables (the LEOPARD cohort was therefore excluded from CAPRA-S comparison).

In the CHIMERA cohort, CAPRA-S achieved a 5-year time-dependent AUC of 0.79, while the multimodal AI-based model demonstrated higher discrimination with a 5-year AUC of 0.82 (ΔAUC = 0.03). When the multimodal AI-derived risk score was combined with CAPRA-S, the integrated model reached a 5-year AUC of 0.83 and showed a significant improvement in model fit (likelihood ratio $\chi^2$ = 8.20, p = 0.004), indicating statistically significant incremental prognostic information beyond guideline-based postoperative risk factors.

In the TCGA-PRAD cohort, CAPRA-S showed higher standalone discrimination than the multimodal AI-based model at 5 years (AUC 0.76 vs. 0.72; ΔAUC = −0.04). Nevertheless, when the multimodal AI-derived risk score was added to CAPRA-S, the combined model achieved a 5-year AUC of 0.79 and showed a significant improvement in model fit compared with CAPRA-S alone (likelihood ratio $\chi^2$ = 8.09, p = 0.004), suggesting that AI-based histopathological features provide complementary prognostic information not fully captured by conventional clinicopathological variables. Detailed comparative performance metrics are reported in Table 2.

Because CAPRA-S relies on postoperative pathological variables unavailable at diagnosis, it was not applicable to the biopsy-based development cohort and was therefore evaluated only in external RP cohorts.

Risk stratification

To assess clinical utility beyond guideline-based postoperative risk stratification, we evaluated the ability of AI-derived risk scores to stratify patients into clinically meaningful risk groups. For LEOPARD, risk stratification was performed using the image-only model, while for CHIMERA and TCGA-PRAD, multimodal model scores were used. Patients were divided into quartiles based on predicted risk, representing progressively higher estimated risk of BCR.

Across all external cohorts, Kaplan–Meier analyses demonstrated a clear and monotonic separation in BCR-free survival across risk quartiles, with patients in higher predicted risk groups experiencing earlier and more frequent recurrence events (Figures 3–5). Differences in survival distributions between the highest- and lowest-risk quartiles were statistically significant in all cohorts (LEOPARD, CHIMERA, and TCGA-PRAD; all p < 0.05). Additional analyses are shown in Supplementary Figures S1–S2, which present Kaplan–Meier risk stratification using the clinical-only model, and in Supplementary Figures S3–S4, which illustrate further stratification of CAPRA-S risk groups by AI-derived quartiles in the CHIMERA and TCGA-PRAD cohorts.

From a clinical perspective, this stratification suggests that AI-derived risk scores can identify subgroups of patients with markedly different recurrence trajectories. Together, these findings demonstrate that AI-based histopathological risk prediction enables meaningful separation of patients into prognostic groups with distinct clinical risk profiles, supporting its potential role as a decision-support tool for postoperative management rather than as a purely statistical predictor.

## Discussion

In this study, we developed and externally validated an AI model to predict BCR after RP using diagnostic prostate biopsy or prostatectomy specimens. The model was trained on preoperative biopsy information from patients undergoing RP in the large, population-based STHLM3 cohort, where it showed strong performance in internal cross-validation. It demonstrated heterogeneous but generally robust performance across three external RP cohorts (LEOPARD, CHIMERA, and TCGA-PRAD) despite differences in patient characteristics, recurrence rates, follow-up durations, histopathology preparation protocols, as well as differences in clinical setting between the training data (preoperative) and validation data (postoperative). This robustness suggests the potential applicability of our approach to diverse clinical settings and patient populations.

The multimodal analysis addresses a broad gap in the literature. In many recent studies, including large multi-institutional works[6,9,23–27], multimodal models combining histopathology-based AI predictions with clinical variables have been shown to outperform image-based AI alone. While this reinforces the complementary value of clinical information, these studies rarely investigate whether clinical variables alone might outperform AI in certain contexts. Our results directly address this question: in TCGA-PRAD, the clinical-only model achieved higher prognostic discrimination, outperforming both the image-only and multimodal AI models. In CHIMERA, the clinical-only model also outperformed the image-only AI model; however, the multimodal AI model surpassed both, indicating that integration of clinical and image-derived features provides complementary prognostic information in this cohort. By showing that clinical-only models can, in some circumstances, match or exceed the performance of image-only AI, our findings underscore the importance of evaluating all three perspectives—image-only, clinical-only, and combined—when assessing prognostic models.

Benchmarking against the guideline-supported CAPRA-S score provided a clinically meaningful reference for real-world postoperative risk stratification. While CAPRA-S demonstrated higher 5-year discrimination than AI alone in TCGA-PRAD, and slightly lower discrimination in CHIMERA, the addition of the multimodal AI-derived risk score significantly improved model fit and discrimination beyond CAPRA-S in both cohorts. This finding indicates that AI-based histopathological features capture prognostic information complementary to standard postoperative clinicopathological variables and are likely to be most useful when combined with established guideline-based risk stratification tools, even in settings where CAPRA-S performs well. In this context, incremental prognostic value may arise from refining risk estimation within clinically defined risk groups rather than from improvements in overall risk ranking alone.

From a clinical perspective, these results suggest that AI-based histopathological risk prediction should be viewed as a complementary tool rather than a replacement for established guideline-based models. In practice, AI-derived risk scores could be integrated alongside existing guideline-based tools (e.g. CAPRA-S) to refine postoperative risk stratification, particularly within intermediate-risk patients where clinical decision-making regarding adjuvant therapy or intensified follow-up remains challenging. Beyond overall discrimination, the risk stratification analyses show how AI scores could be used in practice. Patients in the highest predicted risk group had clearly higher recurrence rates and may benefit from closer surveillance or early consideration of adjuvant therapy, while those in the lowest-risk group had favorable outcomes and could potentially undergo less intensive follow-up. Although quartiles were used here mainly for illustration and comparability across cohorts, these results indicate that clinically meaningful thresholds balancing sensitivity and specificity could be defined in future prospective studies.

In the preoperative setting, biopsy-based AI models may support earlier risk assessment and treatment planning in scenarios where postoperative pathological information is not yet available. The ability to predict BCR using only diagnostic biopsy material is particularly notable, as it suggests that prognostically relevant morphological signals are already present before surgery. This is clinically relevant because most existing postoperative risk tools rely on information unavailable at diagnosis. In addition, our preoperative clinical-only model used only a small set of routinely available variables (age, PSA, and biopsy ISUP grade), and did not include more detailed staging information such as clinical T-stage. Image-based and multimodal models achieved comparable performance even under these constrained preoperative conditions, supporting their potential value for early risk stratification and treatment planning.

Our study has limitations. The external cohorts, particularly CHIMERA, were relatively small, which may affect the stability of performance estimates. Complete clinical data were available for only two of the three external validation cohorts (CHIMERA and TCGA-PRAD), limiting the scope of multimodal analyses. In addition, although the model was trained on pre-treatment biopsy samples, external validation was performed in postoperative RP cohorts. This cross-tissue validation represents a clinically relevant shift in decision context and supports the robustness and generalizability of biopsy-trained models, but prospective validation in fully preoperative settings is still needed to assess clinical utility at the time of diagnosis. Future studies should investigate more advanced multimodal integration strategies beyond simple feature concatenation, explore end-to-end learning approaches, and evaluate alternative foundation models in larger, multi-institutional datasets to better capture complementary information from clinical and image-based inputs.

In conclusion, we demonstrate that an AI model trained on pre-treatment prostate histopathology can provide clinically relevant prognostic information across diverse populations, and that integration with established clinical predictors further improves performance. Crucially, our findings show that clinical variables alone can sometimes outperform image-based AI, reinforcing the value of transparent multimodal evaluation in prognostic modelling, suggesting that the added value of new AI-based prognostication tools should be openly demonstrated through direct comparison with simpler models.

## Methods

Study design and patient cohorts

This study involved the development and validation of an AI model for predicting risk of BCR in individuals with prostate cancer, using hematoxylin and eosin (H&E)-stained slides from RP or diagnostic biopsy specimens. Patients were included if they underwent RP with available digital histopathology and follow-up data.

Four independent cohorts of individuals with localized prostate cancer who underwent RP were included in the study: a biopsy-based cohort (preoperative) from the STHLM3 trial (ISRCTN84445406)[17], two publicly available RP cohorts from Radboud University Medical Center (LEOPARD[15] and CHIMERA[16]), and a well-established RP cohort from The Cancer Genome Atlas (TCGA-PRAD).

For model development, we selected 676 patients from the biopsy-based STHLM3 cohort. These patients were selected from a previously defined subset who underwent RP within one year of diagnosis and achieved postoperative PSA nadir[28], and were further filtered to include only those with available histopathological images. BCR was defined as two consecutive postoperative PSA values ≥0.2 ng/mL. Biopsy slides were digitized using multiple slide scanner models from different scanner manufacturers: Hamamatsu, Philips, Aperio, and Grundium. The distribution of slides across scanners and scanner identifiers is detailed in Supplementary Table S2.

Model validation was performed on three publicly available RP cohorts: LEOPARD ($n_{patients}$=508), CHIMERA ($n_{patients}$=95), and TCGA-PRAD ($n_{patients}$=379). The primary outcome for TCGA-PRAD was BCR, defined as two consecutive postoperative PSA values ≥0.2 ng/mL. For CHIMERA and LEOPARD, BCR was originally defined as a single postoperative PSA value ≥0.1 ng/mL. Histology specimens for TCGA-PRAD were available as WSIs scanned using multiple Aperio ScanScope system at 20x magnification, while slides in the CHIMERA and LEOPARD cohorts were digitized using the 3DHISTECH PANNORAMIC 1000 scanner at 20x magnification.

**Table 1** summarizes the cohorts' characteristics, including number of patients and slides, BCR events, follow-up duration, and selected clinical variables such as age, PSA, ISUP grade group, and CAPRA-S risk group.

## Slide preparation

WSIs were processed to extract representative patch-level features for downstream analysis. During tiling, tissue masks were applied so that only tiles containing tissue were retained. For the LEOPARD and CHIMERA cohorts, we used the tissue masks provided by the challenge organizers. For the TCGA-PRAD and STHLM3 cohorts, tissue regions were segmented using a U-Net-based tissue detection model trained in-house, as described in[29].

WSIs were tiled into fixed-size image patches. In the development cohort (STHLM3), WSIs were tiled into 256 × 256 pixel patches at a resolution of 1.0 µm/pixel (approximately 10x magnification), using a stride of 128 pixels (i.e., 50% overlap), and only tiles containing ≥20% tissue were retained. For the validation cohorts (LEOPARD, CHIMERA, TCGA-PRAD), WSIs were tiled at 10× magnification with a stride of 128 pixels and only tiles containing ≥50% tissue were kept.

## Model details

Following tissue tiling, tile-level feature embeddings were extracted using three state-of-the-art foundation models: UNI2, Virchow2, and CONCH. UNI2 is a vision transformer (ViT) pretrained with self-supervised learning (DINOv2) on over 200 million image tiles sampled from more than 350,000 H&E and IHC slides[18]. Virchow2 is a large-scale ViT (~632 million parameters) trained on 3.1 million WSIs across varied tissues and stain protocols[19]. CONCH is a vision-language foundation model (VLT) pretrained on ~1.17 million histopathology image–caption pairs, combining visual and textual information in its representation learning[20].

Each model generates tile-level embeddings—1536 dimensions for UNI2, 2560 for Virchow2, and 512 for CONCH—serving as input to a MIL framework, which aggregates tile-level features from all available slides of a patient into a single patient-level representation. In addition to training individual pipelines, we also constructed an embedding-level ensemble by concatenating the tile embeddings from all three models into a single feature vector prior to MIL aggregation (4608 dimensions). The aggregated embeddings were then passed through a two-layer multilayer perceptron (MLP) to produce a continuous risk score, optimized using a Cox proportional hazards loss to model time-to-event outcomes.

For the clinical-only model, structured variables (age, PSA, ISUP grade group) were encoded using two fully connected layers that mapped the three input features to a 256-dimensional space and then to a 128-dimensional clinical feature representation. The multimodal model extended the image-only MIL pipeline by concatenating this clinical embedding with the image-derived patient-level features prior to final prediction through an additional two-layer MLP. The overall multimodal architecture is illustrated in Supplementary Figure S5.

All three model types—clinical-only, image-only, and multimodal—shared the same risk prediction architecture after embedding, and were trained independently using identical folds and loss functions.

Further implementation details and hyperparameter configurations are provided in Supplementary Table S1.

## Model development and validation

Model development and validation procedures are illustrated in Figure 6. During development (panel a), we used the STHLM3 cohort for internal nested 5-fold cross-validation. Each fold consisted of training (60%), validation (20%), and held-out test sets (20%). The best-performing model from each fold—based on validation concordance index (C-index)—was retained to construct an ensemble, allowing performance assessment on unseen patients while optimizing hyperparameters.

To ensure robust training, we used stratified 5-fold cross-validation, balancing for BCR status, follow-up time (quantile-binned), and ISUP grade group. Training was conducted for a minimum of 100 and up to 300 epochs, with early stopping applied based on validation C-index. Training was halted if no improvement in performance was observed for 20 consecutive epochs.

To reduce scanner-related bias, since STHLM3 biopsies were digitized using multiple scanner systems from four different vendors (Hamamatsu, Philips, Aperio, and Grundium), we randomly sampled WSIs per patient during training. This increased exposure to scanner variability and limited scanner-specific artifacts.

After model selection, during external validation (Figure 6, panel b), five final models per modality were trained using approximately 80% of the STHLM3 cohort, with early stopping based on the remaining 20%. Each of the five fold-specific models was then applied to LEOPARD, CHIMERA, and TCGA-PRAD, and their predicted risk scores were averaged to obtain ensemble predictions. The STHLM3 cohort was not reused for performance evaluation after this step.

Evaluation metrics and statistical analysis

Model performance during external validation was assessed primarily using the 5-year time-dependent area under the receiver operating characteristic curve (AUC)[30] for BCR risk, reflecting clinically relevant discrimination at a fixed time horizon. The 95% confidence intervals were estimated using nonparametric bootstrapping with 1,000 resamples.

For guideline-based comparison, discrimination of the CAPRA-S score was assessed using the same 5-year time-dependent AUC framework as for the AI-based models. The incremental prognostic value of AI-derived risk scores beyond CAPRA-S was evaluated using likelihood ratio tests comparing nested Cox proportional hazards models with and without inclusion of the AI risk score.

Risk stratification analyses were performed by dividing patients into quartiles based on model-predicted risk scores. Differences in BCR–free survival across risk groups were assessed using Kaplan–Meier estimates and two-sided log-rank tests. Statistical significance was defined as $p < 0.05$.

## Data Availability

This study used four independent datasets comprising WSIs from RP specimens and diagnostic core needle biopsies: LEOPARD (RP): Publicly available as part of the LEOPARD Challenge hosted by Radboud University Medical Center, under the CC BY-NC-SA license. CHIMERA (RP): Publicly available as part of the CHIMERA Challenge hosted by Radboud University Medical Center, under the CC BY-NC-SA 4.0 license. TCGA-PRAD (RP): Publicly available through The Cancer Genome Atlas via the Genomic Data Commons (GDC), licensed under CC BY 3.0. STHLM3 (biopsies): Collected from patients enrolled in the STHLM3 trial who underwent RP within one year of diagnosis. Due to patient privacy regulations and institutional policy, access to this dataset requires a data-sharing agreement and approval from the data access committee at Karolinska Institutet. Details on data licensing and access procedures for public datasets are available via the original sources.

## Code Availability

Core components of our pipeline build on open-source software, including PyTorch (https://github.com/pytorch/pytorch) and publicly available implementations of multiple instance learning and vision transformer architectures. For multiple instance learning, we used the AttentionDeepMIL implementation from AMLab Amsterdam (https://github.com/AMLab-Amsterdam/AttentionDeepMIL). For vision transformer backbones and training utilities, we used the timm library (https://github.com/huggingface/pytorch-image-models). For foundation models, we used publicly available pretrained models including UNI2 (https://huggingface.co/MahmoodLab/UNI), Virchow2 (https://huggingface.co/paige-ai/Virchow2), and CONCH (https://huggingface.co/MahmoodLab/CONCH). Due to institutional intellectual property policies and ongoing research commercialization activities, the full codebase cannot currently be released. However, all steps of model design, training, and evaluation are described in sufficient detail in the Methods and Supplementary Materials to enable independent replication. Reasonable requests for further technical details may be considered by the corresponding author.

## Acknowledgements

We gratefully acknowledge the STHLM3 study team for their contribution to patient recruitment, data collection, and clinical coordination, which made this study possible. We also thank the patients who participated in the STHLM3 study and generously contributed their clinical data. We also acknowledge The Cancer Genome Atlas Prostate Adenocarcinoma (TCGA-PRAD) project and the CHIMERA and LEOPARD challenge organizers for providing data used in this study, as well as the patients and investigators whose contributions made these resources available. M.E. received funding from the Swedish Research Council, Swedish Cancer Society, Swedish Prostate Cancer Society, Nordic Cancer Union, Karolinska Institutet, and Region Stockholm. K.K. received funding from the SciLifeLab & Wallenberg Data Driven Life Science Program (KAW 2024.0159), Swedish Cancer Society, Instrumentarium Science Foundation and Karolinska Institutet Research Foundation. High-performance computing was supported by the National Academic Infrastructure for Supercomputing in Sweden (NAISS) and the Swedish National Infrastructure for Computing (SNIC) at C3SE, partially funded by the Swedish Research Council through grant agreements no. 2022-06725 and no. 2018-05973, and by the Berzelius supercomputing resource provided by the National Supercomputer Centre at Linköping University and the Knut and Alice Wallenberg Foundation.

# Ethics declarations

## Competing interests

A.C. has received honoraria from Astellas Pharma AB. N.M., K.K., and L.E. are shareholders of Clinsight AB. M.E. is a stockholder of Clinsight AB and A3P Biomedical AB, and holds patents in A3P Biomedical AB. T.N. owns stock in A3P Biomedical AB. H.G. holds patents and owns stock in A3P Biomedical AB. C.M. and T.P. are employed by A3P Biomedical AB.

## Ethical considerations

This study includes data from the Swedish STHLM3 cohort. All data were de-identified at the source and transferred to Karolinska Institutet in anonymized form. The study was conducted in accordance with the Declaration of Helsinki and approved by the Stockholm Regional Ethics Committee (permits 2012/572-31/1 and 2012/438-31/3) and the Swedish Ethical Review Authority (permit 2019-05220). Informed consent was obtained from all participants in the Swedish cohort. The external cohorts used for validation are publicly available and were analyzed in accordance with the data access and usage policies of their respective repositories.

# Figures and Tables

## Tables

|  | STHLM3 | LEOPARD | CHIMERA | TCGA-PRAD |
|---|---|---|---|---|
| Tissue specimen | Biopsy | Prostatectomy | Prostatectomy | Prostatectomy |
| Number of patients | 676 | 508 | 95 | 379 |
| Number of slides | 7366 | 508 | 190 | 877 |
| Age at diagnosis (years) | 64.3 (59.3-67.3) | - | 66.0 (60.0-69.0) | 61.0 (56.0-66.0) |
| Pre-treatment PSA (ng/mL) | 4.3 (3.2-6.6) | - | 7.8 (5.2-12.0) | 7.4 (5.2-10.8) |
| ISUP | | | | |
| 1 | 120 (17.8%) | - | 8 (8.4%) | 41 (10.6%) |
| 2 | 346 (51.2%) | - | 44 (46.3%) | 114 (30.1%) |
| 3 | 113 (16.7%) | - | 27 (28.4%) | 72 (19.0%) |
| 4 | 0 (0.0%) | - | 7 (7.4%) | 62 (16.4%) |
| 5 | 97 (14.3%) | - | 9 (9.5%) | 90 (23.7%) |
| CAPRA-S | | | | |
| Low (0-2) | - | - | 20 (21.1%) | 161 (42.5%) |
| Intermediate (3-5) | - | - | 54 (56.8%) | 131 (34.6%) |
| High (6-12) | - | - | 21 (22.1%) | 87 (23.0%) |
| Number of BCR events | 183 | 87 | 27 | 48 |
| Median follow-up (years) | 9.0 (7.3-9.4) | 4.8 (1.2-7.4) | 3.2 (1.6-4.9) | 2.3 (1.2-3.8) |
| Median time to BCR (years) | 4.3 (2.4-6.6) | 4.2 (1.3-7.6) | 0.4 (0.1-1.6) | 1.6 (0.6-2.7) |

**Table 1.** Clinicopathological characteristics of the study cohorts. Summary of tissue source, sample size, number of slides, BCR events, follow-up times, patient age, pre-treatment PSA, ISUP grade group, and CAPRA-S distribution for the STHLM3, LEOPARD, CHIMERA, and TCGA-PRAD cohorts. Continuous variables are reported as median (interquartile range, IQR). Categorical variables are reported as counts with percentages in parentheses, calculated within each cohort.

| Cohort | 5 year AUC | | ΔAUC | 5 year AUC | Likelihood ratio test | |
|---|---|---|---|---|---|---|
| | Multimodal AI | CAPRA-S | | Multimodal AI + CAPRA-S | χ² | p-value |
| CHIMERA | 0.82 [0.69-0.94] | 0.79 [0.67-0.89] | +0.03 [-0.08-0.14] | 0.83 [0.71-0.95] | 8.20 | 0.004 |
| TCGA-PRAD | 0.72 [0.61-0.82] | 0.76 [0.67-0.84] | −0.04 [-0.14-0.05] | 0.79 [0.69-0.88] | 8.09 | 0.004 |

**Table 2.** Comparison of AI-based risk prediction with CAPRA-S in external RP cohorts. Time-dependent AUCs at 5 years are reported for the multimodal AI model, CAPRA-S, and the combined multimodal AI + CAPRA-S model in the CHIMERA and TCGA-PRAD cohorts. ΔAUC represents the difference in 5-year AUC between the multimodal AI model and CAPRA-S. Values are shown with 95% confidence intervals. Likelihood ratio tests ($\chi^2$ and p-values) evaluate the incremental prognostic value of adding the AI-derived risk score to CAPRA-S.

Figures

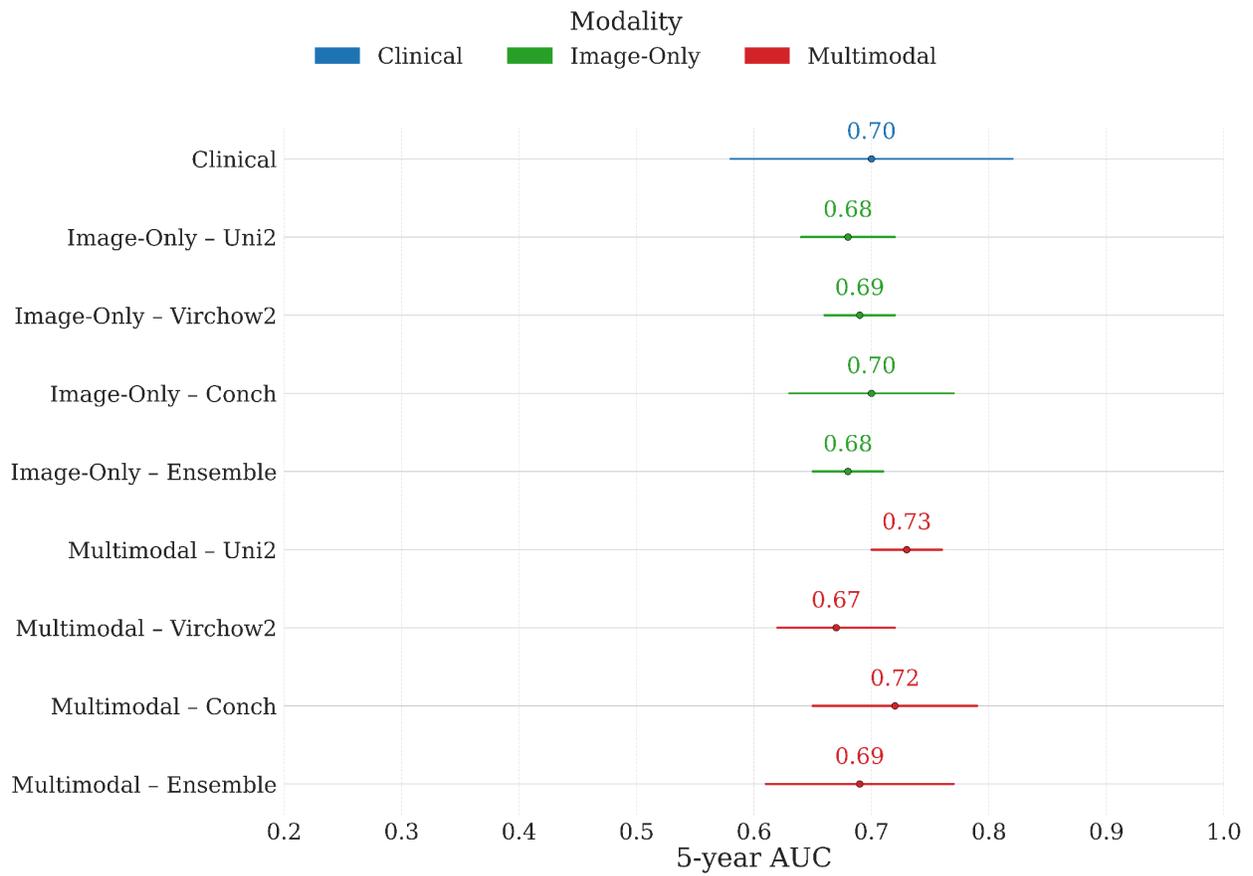

**Figure 1:** Performance of clinical-only, image-only, and multimodal models reported as mean ± standard deviation across nested 5-fold cross-validation on STHLM3. Time-dependent AUC at 5 years. Image-only and multimodal models were developed using three tile-level encoders (UNI2, Virchow2, CONCH) and their encoder-level ensemble.

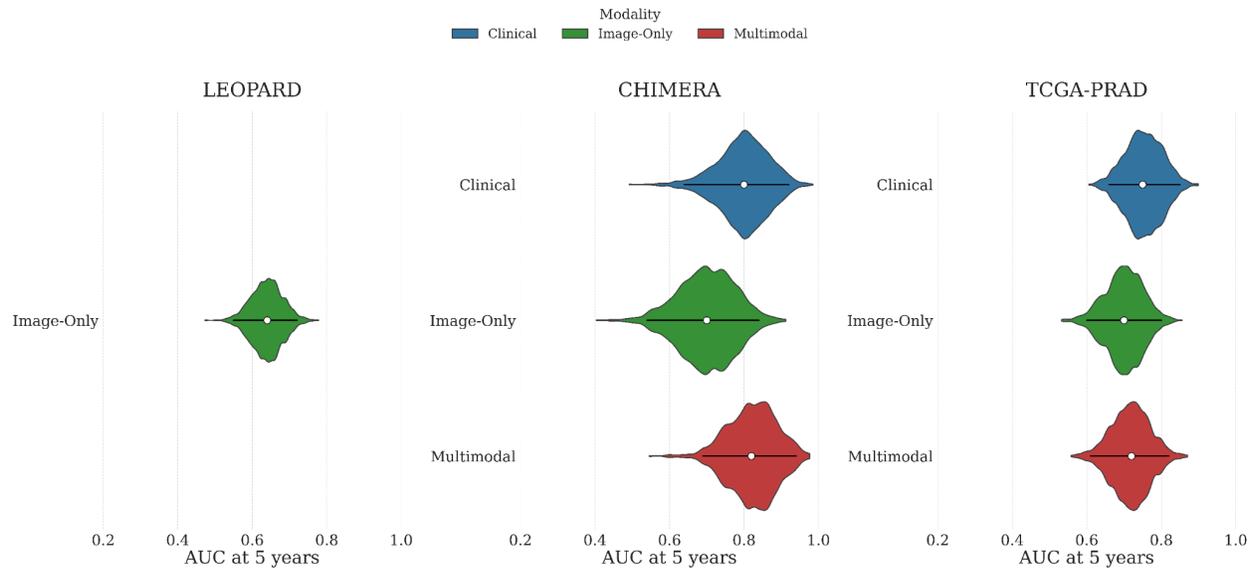

**Figure 2:** External validation performance of the final selected models across the three external RP cohorts. Violin plots show the distribution of bootstrapped time-dependent AUC at 5 years values for each modality: clinical-only (blue), image-only (green, CONCH), and multimodal (red, UNI2 + clinical). In LEOPARD, only image data were available; CHIMERA and TCGA-PRAD included both image and clinical data, enabling evaluation of all three modalities. The white dot indicates the mean AUC, and the height reflects the density of bootstrap replicates.

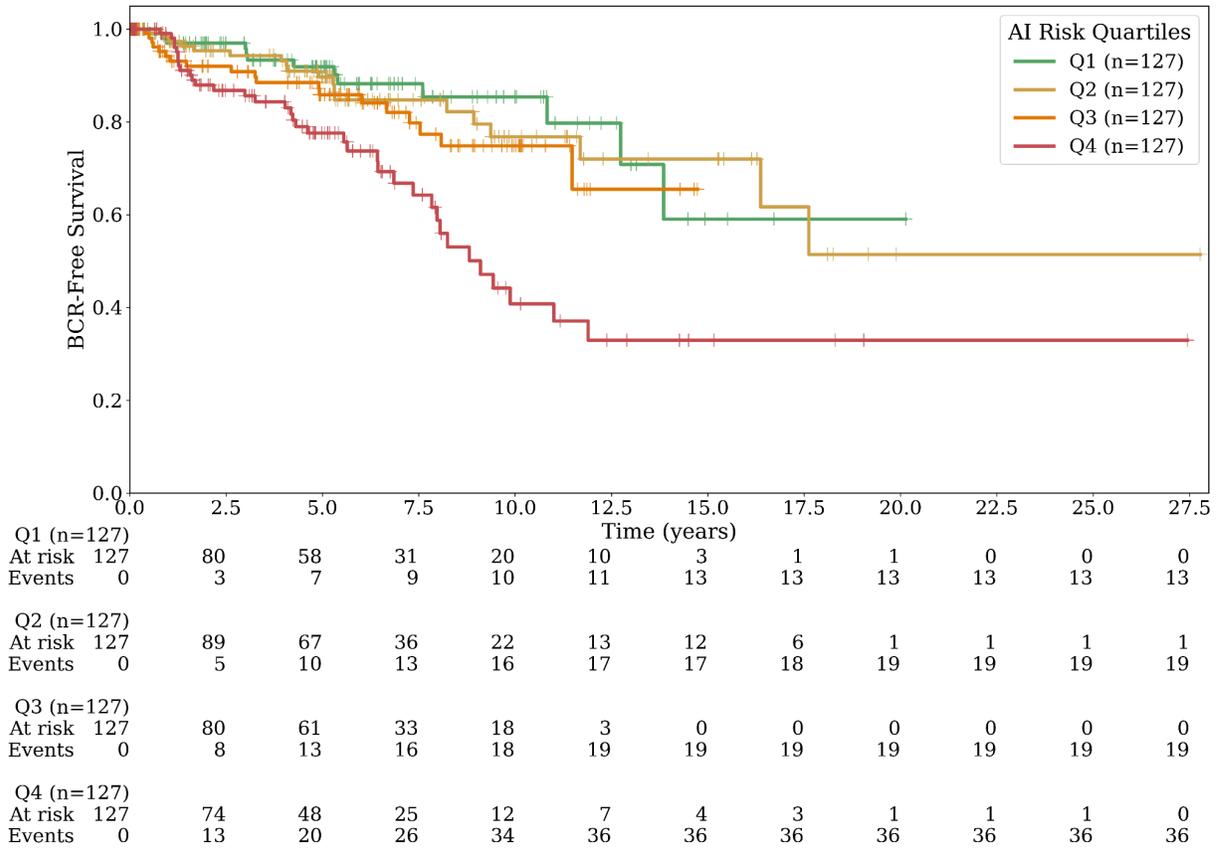

**Figure 3. Kaplan–Meier survival curves stratified by AI-predicted risk quartiles in the LEOPARD cohort.** Patients were divided into quartiles based on their model-predicted risk of BCR. BCR-free survival decreased progressively across quartiles, with the highest-risk group (Q4) showing significantly worse outcomes compared to the lowest-risk group (Q1). The log-rank test confirmed significant differences in survival distributions ($p < 0.05$).

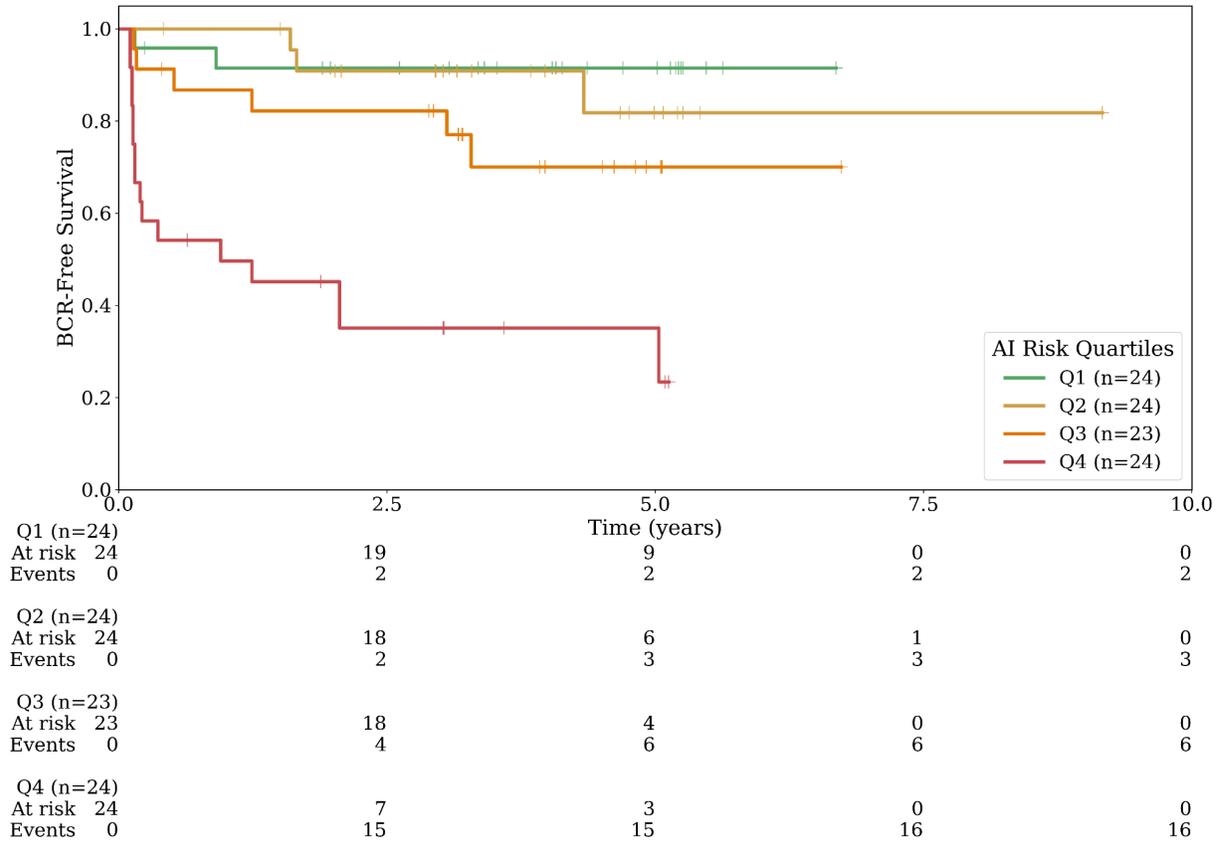

**Figure 4. Kaplan–Meier survival curves stratified by AI-predicted risk quartiles in the CHIMERA cohort.** Patients were divided into quartiles based on their model-predicted risk of BCR. BCR-free survival decreased progressively across quartiles, with the highest-risk group (Q4) showing significantly worse outcomes compared to the lowest-risk group (Q1). The log-rank test confirmed significant differences in survival distributions ($p < 0.05$).

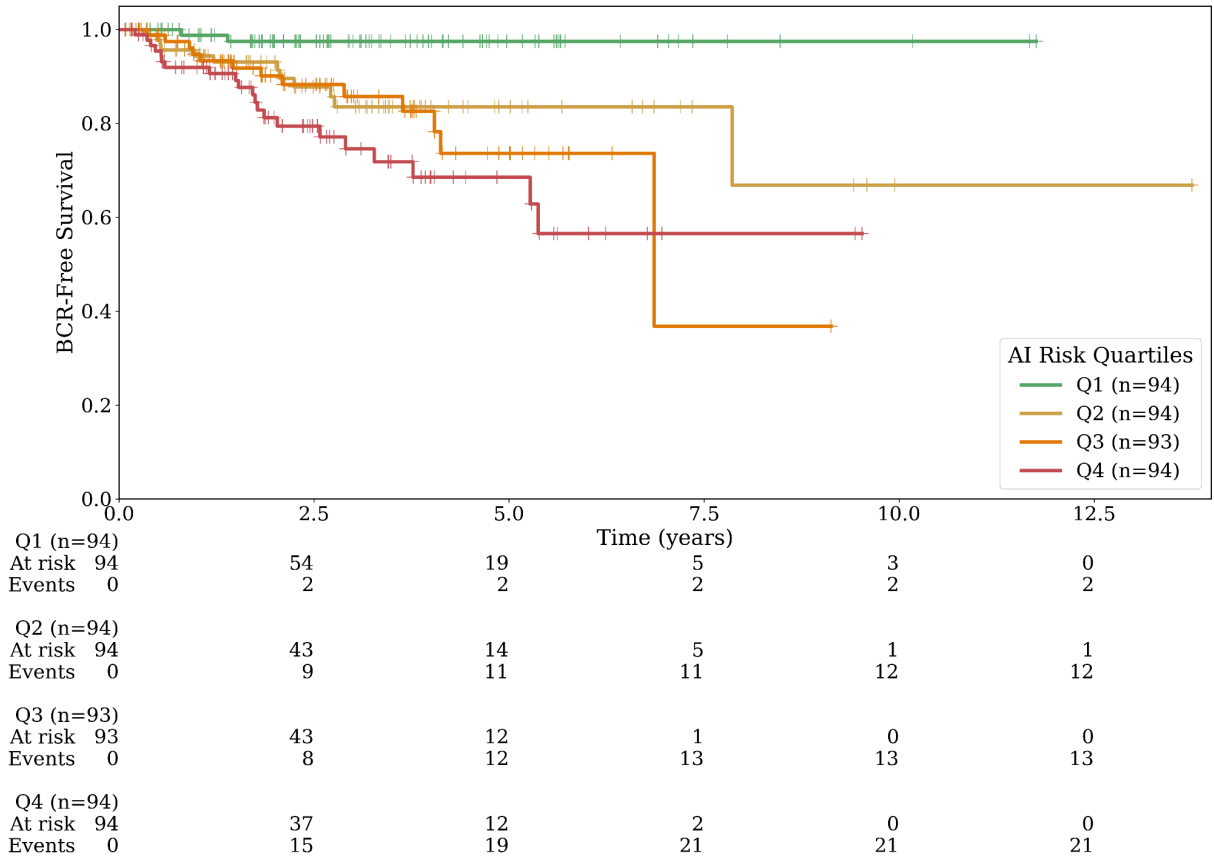

**Figure 5. Kaplan–Meier survival curves stratified by AI-predicted risk quartiles in the TCGA-PRAD cohort.** Patients were divided into quartiles based on their model-predicted risk of BCR. BCR-free survival decreased progressively across quartiles, with the highest-risk group (Q4) showing significantly worse outcomes compared to the lowest-risk group (Q1). The log-rank test confirmed significant differences in survival distributions ($p < 0.05$).

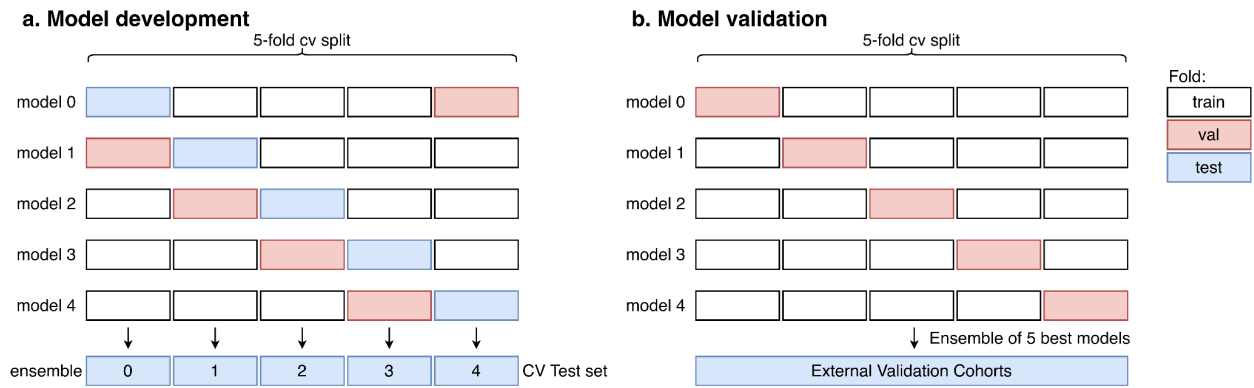

**Figure 6**. **(a)** Model development: training and validation using nested 5-fold cross-validation. The best-performing configuration in each fold was selected based on performance on the internal validation (test) set. **(b)** Model validation: the five best cross-validated models were retrained on the full STHLM3 dataset and evaluated on external validation cohorts.

# Supplementary Materials

## A. Figures and Tables

| Component | Description |
|---|---|
| Modalities | Clinical-only, Image-Only, Multimodal |
| Clinical input | Age at diagnosis, pre-treatment PSA, ISUP grade group |
| Clinical normalization | All clinical variables were standardized using Z-score normalization |
| Image input | Tile-level embeddings extracted from WSIs using UNI2 (1536-d), Virchow2 (2560-d), or CONCH (512-d) |
| Multimodal input | Concatenation of clinical features with patient-level image representation |
| Max tiles per bag | 3500 |
| Batch size | 256 bags |
| MIL architecture | Attention-based Multiple Instance Learning (MIL) |
| Final activation | None (raw risk score, used with Cox loss) |
| Loss function | Cox proportional hazards loss |
| Optimizer | Adam |
| Learning rate | 1e-4 |
| Gradient accumulation | 1 step |
| Early stopping | Patience = 20 epochs (based on C-index) |

| | | |
|---|---|
| Epochs | Min: 100, Max: 300 |
| CV strategy | Nested 5-fold (stratified by ISUP, BCR, follow-up) |

**Supplementary Table S1. Hyperparameters for Model Training and Evaluation.**

| | STHLM3RP | LEOPARD | CHIMERA | TCGA-PRAD |
|---|---|---|---|---|
| Number of Patients | 676 | 508 | 95 | 379 |
| Number of Slides | 7366 | 508 | 190 | 877 |
| Number of WSIs | 15162 | 508 | 190 | 877 |
| Hamamatsu (760347) | 3023 (19.9%) | 0 (0.0%) | 0 (0.0%) | 0 (0.0%) |
| Aperio (RUD-D10971) | 2487 (16.4%) | 0 (0.0%) | 0 (0.0%) | 0 (0.0%) |
| Hamamatsu (870003) | 2718 (17.9%) | 0 (0.0%) | 0 (0.0%) | 0 (0.0%) |
| Philips (FMT0047) | 5912 (39.0%) | 0 (0.0%) | 0 (0.0%) | 0 (0.0%) |
| Grundium (MGU-00003-000184) | 1022 (6.7%) | 0 (0.0%) | 0 (0.0%) | 0 (0.0%) |
| 3DHISTECH (PANNORAMIC 1000) | 0 (0.0%) | 508 (100.0%) | 190 (100.0%) | 0 (0.0%) |
| Aperio (SS1764CNTLR) | 0 (0.0%) | 0 (0.0%) | 0 (0.0%) | 182 (20.8%) |
| Aperio (SS1552CNTLR) | 0 (0.0%) | 0 (0.0%) | 0 (0.0%) | 46 (5.2%) |
| Aperio (SS1302) | 0 (0.0%) | 0 (0.0%) | 0 (0.0%) | 313 (35.7%) |
| Aperio (SS1763CNTLR) | 0 (0.0%) | 0 (0.0%) | 0 (0.0%) | 84 (9.6%) |
| Aperio (SS1641) | 0 (0.0%) | 0 (0.0%) | 0 (0.0%) | 21 (2.4%) |
| Aperio (SS1248CNTLR) | 0 (0.0%) | 0 (0.0%) | 0 (0.0%) | 14 (1.6%) |
| Aperio (SS1511CNTLR) | 0 (0.0%) | 0 (0.0%) | 0 (0.0%) | 54 (6.2%) |
| Aperio (SS1546) | 0 (0.0%) | 0 (0.0%) | 0 (0.0%) | 56 (6.4%) |
| Aperio (SS1534CNTLR) | 0 (0.0%) | 0 (0.0%) | 0 (0.0%) | 20 (2.3%) |
| Aperio (SS1436CNTLR) | 0 (0.0%) | 0 (0.0%) | 0 (0.0%) | 34 (3.9%) |
| Aperio (SS1289) | 0 (0.0%) | 0 (0.0%) | 0 (0.0%) | 13 (1.5%) |
| Aperio (Unknown) | 0 (0.0%) | 0 (0.0%) | 0 (0.0%) | 23 (2.6%) |
| Aperio (SS1540CNTLR) | 0 (0.0%) | 0 (0.0%) | 0 (0.0%) | 9 (1.0%) |
| Aperio (SS7168CNTLR) | 0 (0.0%) | 0 (0.0%) | 0 (0.0%) | 1 (0.1%) |
| Aperio (SS1475) | 0 (0.0%) | 0 (0.0%) | 0 (0.0%) | 7 (0.8%) |

**Supplementary Table S2. Number of patients, slides, WSIs, and scanner types per cohort.** Scanner entries include both manufacturer and scanner serial number (shown in parentheses). STHLM3 used

multiple scanners from different manufacturers, LEOPARD and CHIMERA used the 3DHISTECH PANNORAMIC 1000 system, and TCGA-PRAD used Aperio scanners with several different scanner identifiers. Percentages indicate the proportion of WSIs scanned with each scanner within each cohort.

| Modality | Encoder | C-Index (mean) | C-Index (std) | 5-year AUC (mean) | 5-year AUC (std) |
|---|---|---|---|---|---|
| **Clinical-Only** | | 0.70 | 0.12 | 0.70 | 0.12 |
| **Image-Only** | UNI2 | 0.68 | 0.04 | 0.68 | 0.04 |
| | Virchow2 | 0.69 | 0.04 | 0.69 | 0.03 |
| | CONCH | 0.68 | 0.09 | 0.70 | 0.07 |
| | Ensemble | 0.68 | 0.03 | 0.68 | 0.03 |
| **Multimodal** | UNI2 | 0.72 | 0.04 | 0.73 | 0.03 |
| | Virchow2 | 0.66 | 0.03 | 0.67 | 0.05 |
| | CONCH | 0.71 | 0.08 | 0.72 | 0.07 |
| | Ensemble | 0.68 | 0.07 | 0.69 | 0.08 |

**Supplementary Table S3. Performance of clinical-only, image-only, and multimodal models reported as mean ± standard deviation across nested 5-fold cross-validation.** Metrics include concordance index (C-index) and time-dependent AUC at 5 years. Image-only and multimodal models were developed using three tile-level encoders (UNI2, Virchow2, CONCH) and their encoder-level ensemble.

| Cohort | Modality | Encoder | C-Index | C-index [95% CI] | 5year AUC | AUC [95% CI] |
|---|---|---|---|---|---|---|
| **LEOPARD** | Image-Only | UNI2 | 0.60 | [0.53, 0.67] | 0.62 | [0.53, 0.70] |
| | | Virchow2 | 0.63 | [0.56, 0.69] | 0.65 | [0.57, 0.74] |
| | | CONCH | 0.63 | [0.57, 0.69] | 0.64 | [0.55, 0.72] |
| | | Ensemble | 0.67 | [0.60, 0.72] | 0.69 | [0.61, 0.77] |
| **CHIMERA** | Clinical-Only | | 0.73 | [0.60, 0.84] | 0.80 | [0.64, 0.92] |
| | Image-Only | UNI2 | 0.78 | [0.67, 0.88] | 0.78 | [0.63, 0.92] |
| | | Virchow2 | 0.65 | [0.51, 0.77] | 0.66 | [0.50, 0.81] |
|  |  | CONCH | 0.68 | [0.55, 0.79] | 0.70 | [0.54, 0.84] |
|  |  | Ensemble | 0.72 | [0.59, 0.82] | 0.75 | [0.60, 0.88] |
|  | Multimodal | UNI2 | 0.79 | [0.68, 0.88] | 0.82 | [0.69, 0.94] |
|  |  | Virchow2 | 0.69 | [0.55, 0.79] | 0.68 | [0.51, 0.82] |
|  |  | CONCH | 0.68 | [0.54, 0.80] | 0.71 | [0.56, 0.87] |
|  |  | Ensemble | 0.74 | [0.62, 0.84] | 0.76 | [0.60, 0.89] |
| TCGA-PRAD | Clinical |  | 0.72 | [0.64, 0.79] | 0.76 | [0.66, 0.85] |
|  | Image-Only | UNI2 | 0.64 | [0.55, 0.73] | 0.66 | [0.55, 0.76] |
|  |  | Virchow2 | 0.52 | [0.41, 0.63] | 0.57 | [0.44, 0.69] |
|  |  | CONCH | 0.67 | [0.58, 0.74] | 0.70 | [0.60, 0.80] |
|  |  | Ensemble | 0.60 | [0.50, 0.70] | 0.64 | [0.53, 0.75] |
|  | Multimodal | UNI2 | 0.70 | [0.61, 0.77] | 0.72 | [0.61, 0.82] |
|  |  | Virchow2 | 0.53 | [0.42, 0.64] | 0.58 | [0.45, 0.69] |
|  |  | CONCH | 0.69 | [0.61, 0.76] | 0.72 | [0.63, 0.82] |
|  |  | Ensemble | 0.62 | [0.53, 0.72] | 0.65 | [0.54, 0.74] |

**Supplementary Table S4. Summary of model performance across external validation cohorts and modalities.** The table reports the time-dependent AUC at 5 years and the concordance index (C-index) computed over the full follow-up period. Results are stratified by cohort, modality (clinical-only, image-only, and multimodal), and foundation model encoder (UNI2, Virchow2, CONCH, or ensemble). For each metric, the point estimate and 95% confidence intervals based on bootstrapping are provided.

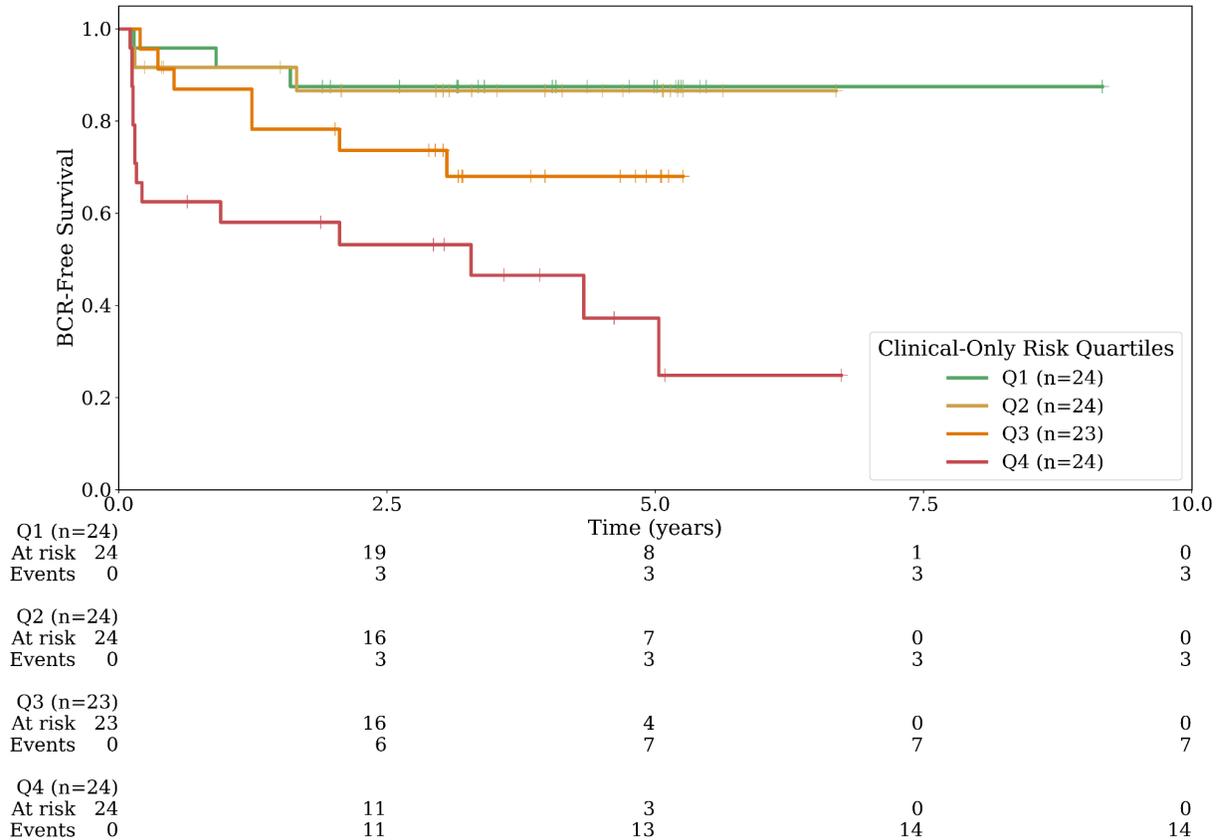

| | | | | |
|---|---|---|---|---|
| Q1 (n=24) | | | | |
| At risk 24 | 19 | 8 | 1 | 0 |
| Events 0 | 3 | 3 | 3 | 3 |
| Q2 (n=24) | | | | |
| At risk 24 | 16 | 7 | 0 | 0 |
| Events 0 | 3 | 3 | 3 | 3 |
| Q3 (n=23) | | | | |
| At risk 23 | 16 | 4 | 0 | 0 |
| Events 0 | 6 | 7 | 7 | 7 |
| Q4 (n=24) | | | | |
| At risk 24 | 11 | 3 | 0 | 0 |
| Events 0 | 11 | 13 | 14 | 14 |

**Supplementary Figure S1. Kaplan–Meier survival curves stratified by clinical-only predicted risk quartiles in the CHIMERA cohort.** Patients were divided into quartiles based on their model-predicted risk of BCR. BCR-free survival decreased progressively across quartiles, with the highest-risk group (Q4) showing significantly worse outcomes compared to the lowest-risk group (Q1). The log-rank test confirmed significant differences in survival distributions ($p < 0.05$).

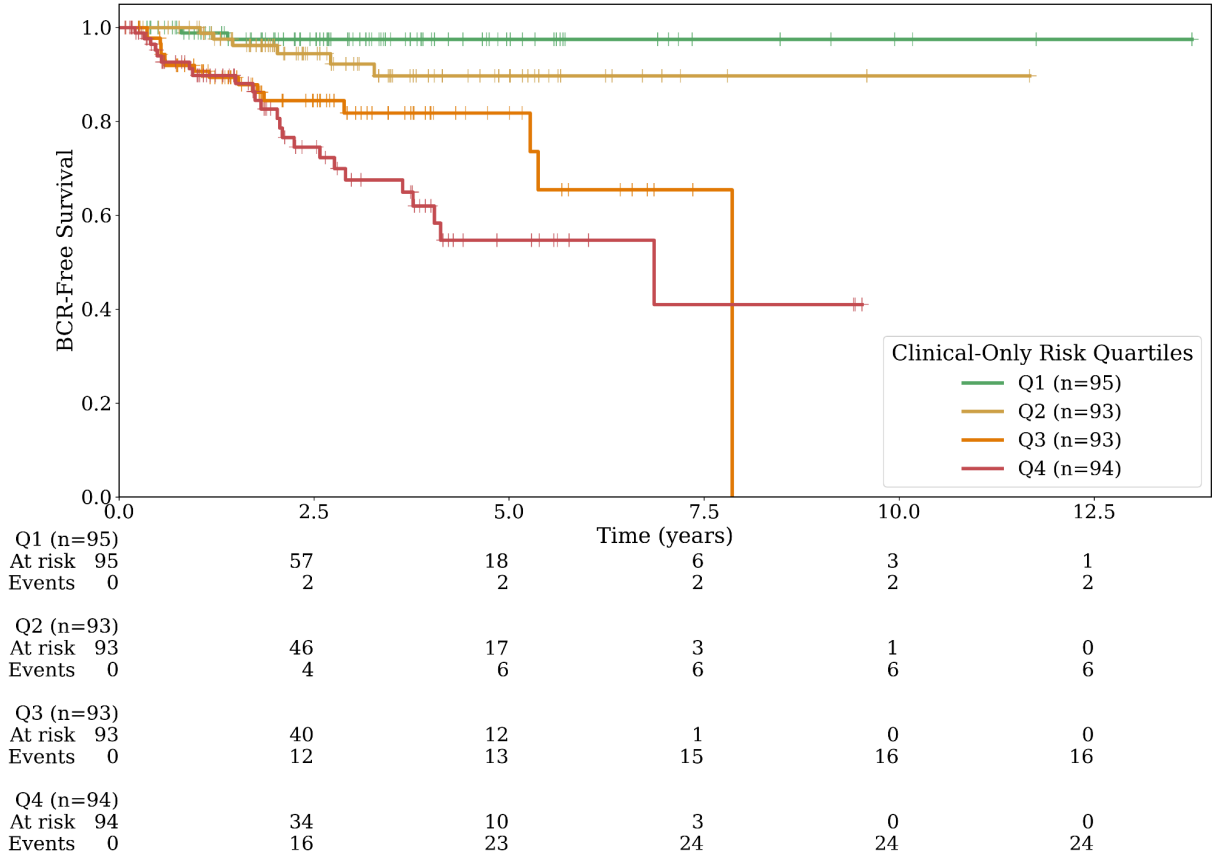

**Supplementary Figure S2. Kaplan–Meier survival curves stratified by clinical-only predicted risk quartiles in the TCGA-PRAD cohort.** Patients were divided into quartiles based on their model-predicted risk of BCR. BCR-free survival decreased progressively across quartiles, with the highest-risk group (Q4) showing significantly worse outcomes compared to the lowest-risk group (Q1). The log-rank test confirmed significant differences in survival distributions (p < 0.05).

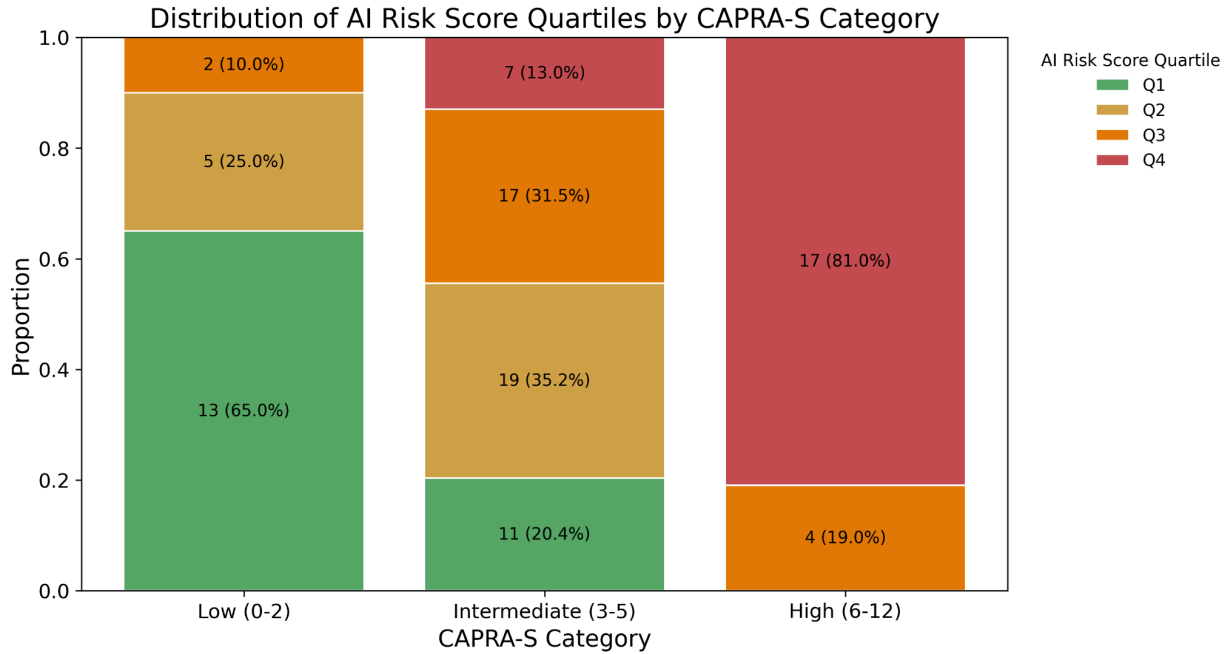

**Supplementary Figure S3. Distribution of multimodal AI risk score quartiles across CAPRA-S risk categories in the CHIMERA cohort.** Each bar represents a CAPRA-S category (low, intermediate, high) and is normalized to 100%, showing the proportion of patients in each AI risk score quartile. Numbers inside the bars indicate the absolute count and corresponding percentage within each CAPRA-S category. This visualization highlights how AI-derived risk stratification varies across established clinical risk groups.

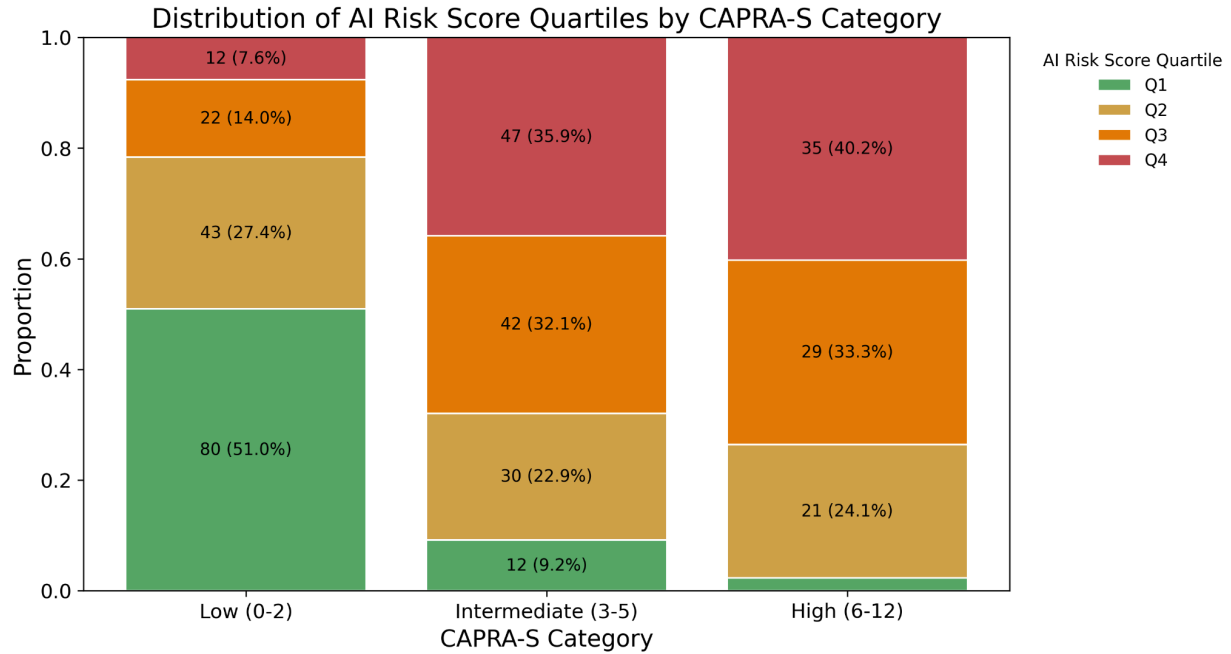

**Supplementary Figure S4. Distribution of multimodal AI risk score quartiles across CAPRA-S risk categories in the TCGA-PRAD cohort.** Each bar represents a CAPRA-S category (low, intermediate, high) and is normalized to 100%, showing the proportion of patients in each AI risk score quartile. Numbers inside the bars indicate the absolute count and corresponding percentage within each CAPRA-S category. This visualization highlights how AI-derived risk stratification varies across established clinical risk groups.

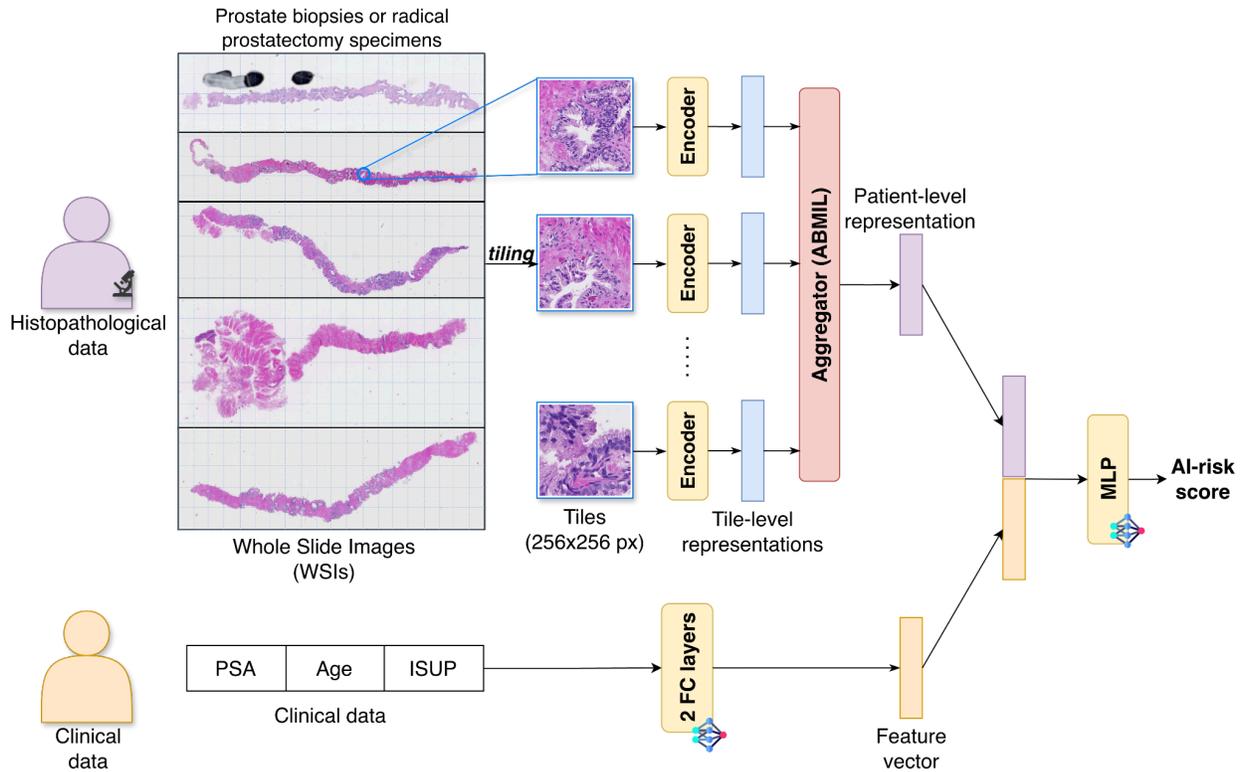

**Supplementary Figure S5. Overview of the multimodal AI pipeline for predicting BCR from prostate histopathology and clinical data.** Whole-slide images (WSIs) are tiled and encoded into feature embeddings using a foundation model. Tile-level embeddings are aggregated into a patient-level representation via attention-based MIL. Structured clinical variables (age, PSA, ISUP) are processed through two fully connected (FC) layers. In the multimodal model, clinical and image representations are concatenated and passed through a two-layer multilayer perceptron (MLP) to generate a continuous risk score optimized with a Cox proportional hazards loss.